\documentclass{article} 
\usepackage{iclr2025_conference,times}
\usepackage{etoolbox}
\usepackage{amsmath}
\usepackage{nicefrac}
\usepackage{graphicx}
\usepackage{tabularx}
\usepackage{subcaption}
\usepackage{times}
\usepackage{latexsym}
\usepackage{dsfont}
\usepackage{graphics}
\usepackage{graphicx}
\usepackage{color}
\usepackage{colortbl}
\usepackage{amsfonts}
\usepackage{amssymb}
\usepackage{bbm}
\usepackage{booktabs}
\usepackage{microtype}
\usepackage{multirow}
\usepackage{multicol}
\usepackage{tabu}
\usepackage{amsthm}
\usepackage{footmisc}
\usepackage{soul}
\usepackage{arydshln}
\usepackage{wasysym}
\usepackage{array}
\usepackage{makecell}
\usepackage{scalerel}
\usepackage{algpseudocode}
\usepackage{xspace}
\usepackage{utfsym}
\usepackage{dirtytalk}
\usepackage{mathtools}
\usepackage{float}
\usepackage{enumitem}
\usepackage{tikz}
\usepackage{tcolorbox}
\usepackage{siunitx}
\usepackage[ruled,vlined]{algorithm2e}
\usepackage{ragged2e}

\usepackage{amsmath,amsfonts,bm}

\def\eqref#1{equation~\ref{#1}}

\def\1{\bm{1}}

\DeclareMathAlphabet{\mathsfit}{\encodingdefault}{\sfdefault}{m}{sl}
\SetMathAlphabet{\mathsfit}{bold}{\encodingdefault}{\sfdefault}{bx}{n}

\providetoggle{showcomments}
\settoggle{showcomments}{true}

\newcolumntype{C}[1]{>{\centering\let\newline\\\arraybackslash\hspace{0pt}}m{#1}}

\DeclarePairedDelimiterX{\infdivx}[2]{(}{)}{%
  #1\;\delimsize\|\;#2%
}

\newtcolorbox[auto counter]{promptfloat}[2][]{%
    float=!hb,%
    blend before title=dash hang,%
    title={\textbf{Prompt~\thetcbcounter:} #2},%
    colback=black!5!white,%
    colframe=black!70!white,%
    #1}
\newcommand{\prompt}[3]{%
    \begin{promptfloat}[label={#1}]{#2}
    {\vspace{8pt}\small \texttt{#3}\vspace{8pt}}
\end{promptfloat}
}

\newcommand*\samethanks[1][\value{footnote}]{\footnotemark[#1]}

\usepackage{hyperref}
\usepackage{url}

\title{Enhancing LLM Robustness to Perturbed\\Instructions: An Empirical Study}

\author{Aryan Agrawal\thanks{Equal contribution. Correspondence to contact@aryanagrawal.com, lisa.alazraki20@imperial.ac.uk.}~, Lisa Alazraki\samethanks~, Shahin Honarvar, Marek Rei \\
Imperial College London
}

\iclrfinalcopy 
\begin{document}

\maketitle

\begin{abstract}
Large Language Models (LLMs) are highly vulnerable to input perturbations, as even a small prompt change may result in a substantially different output. Existing methods to enhance LLM robustness are primarily focused on perturbed data samples, whereas improving resiliency to perturbations of task-level instructions has remained relatively underexplored. In this work, we focus on character- and word-level edits of task-specific instructions, which substantially degrade downstream performance. We experiment with a variety of techniques to enhance the robustness of LLMs, including self-denoising and representation alignment, testing different models (Llama 3 and Flan-T5), datasets (CoLA, QNLI, SST-2) and instructions (both task-oriented and role-oriented). We find that, on average, self-denoising—whether performed by a frozen LLM or a fine-tuned model—achieves substantially higher performance gains than alternative strategies, including more complex baselines such as ensembling and supervised methods. We share our data and code at \href{https://github.com/ary4n99/llm-robustness}{\texttt{https://github.com/ary4n99/llm-robustness}}.
\end{abstract}


\section{Introduction}

Despite achieving impressive performance in increasingly sophisticated tasks \citep{nori2023capabilitiesgpt4medicalchallenge, roemmele-gordon-2024-test}, LLMs remain sensitive to input perturbations \citep{DBLP:journals/debu/0001HH0ZWY0HGJ024}. While human performance in natural language tasks is resilient to small alterations in the problem description \citep{walkington2019}, LLMs have consistently been observed to shift their output dramatically even with minor changes to the input \citep{moradi-samwald-2021-evaluating, wang-etal-2022-measure, pmlr-v239-alazraki23a, gulati2024putnamaxiom, honarvar2025turbulencesystematicallyautomaticallytesting}. Prior literature extensively investigates improving LLM robustness when individual data samples are perturbed \citep{10.1145/3637528.3671932, wang-etal-2024-resilience, chen2025secaligndefendingpromptinjection}. On the other hand, methods for handling perturbations of task-level instructions (i.e., fixed templates that are combined with each data point to help solve a task) are less researched.

Intuitively, instruction quality and readability have a considerable impact on performance: perturbing the instruction can potentially lead the LLM to misunderstand the task and fail on all samples. Indeed, prior work finds that LLM proficiency varies widely when instructions are paraphrased \citep{mizrahi-etal-2024-state, DBLP:journals/jmlr/ZhuZ00024} or individual words are replaced, added or removed \citep{gu-etal-2023-robustness, DBLP:journals/jmlr/ZhuZ00024}. Similarly, \citet{sun2024evaluating} 
study the performance of instruction-tuned LLMs when test-time instructions are phrased differently from the training data, and observe substantial degradation across different models and tasks. As a solution, they propose aligning the internal model representations of the rephrased instructions to those of the original ones.

In this work, we investigate a range of methods---both prompt-based and fine-tuned---for enhancing LLM robustness to perturbed instructions in classification tasks. We focus on word- and character-level perturbations, as these have been found to cause the greatest performance decline \citep{DBLP:journals/jmlr/ZhuZ00024}. We assess each method on a combination of six instructions, two types of perturbations, three datasets, and two base models. Our experiments show that LLMs are particularly effective at self-denoising instructions, especially when the process is done iteratively.

Our main findings are as follows: (1) iterative self-denoising---whether carried out by a fine-tuned model or the base LLM---prevents a considerable portion of the performance drop caused by using perturbed instructions in classification tasks; (2) self-denoising in general is far more effective than other methods including instruction ensembling and hidden representation alignment; (3) other denoising strategies, such as perplexity smoothing, are not as successful. In fact, they tend to decrease performance further.


\section{Methods}
\label{sec:methods}

\begin{figure}[!t]
\begin{subfigure}{0.33\textwidth}
\includegraphics[width=\textwidth]{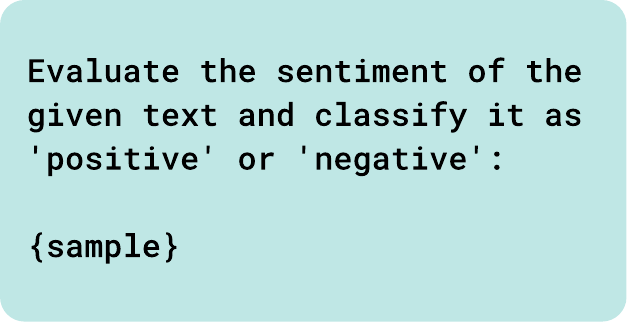}
\caption{Non-perturbed.}
\label{fig:nonperturbed}
\end{subfigure}
\begin{subfigure}{0.33\textwidth}
\includegraphics[width=\textwidth]{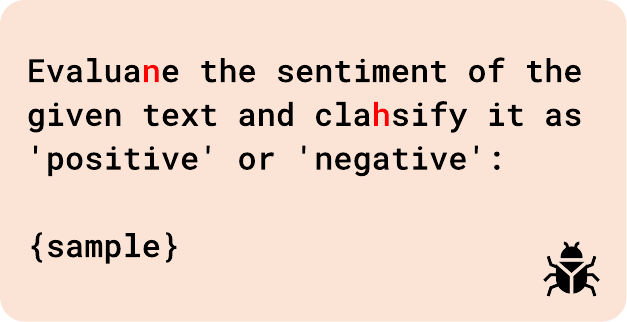}
\caption{DeepWordBug.}
\label{fig:character}
\end{subfigure}
\begin{subfigure}{0.33\textwidth}
\includegraphics[width=\textwidth]{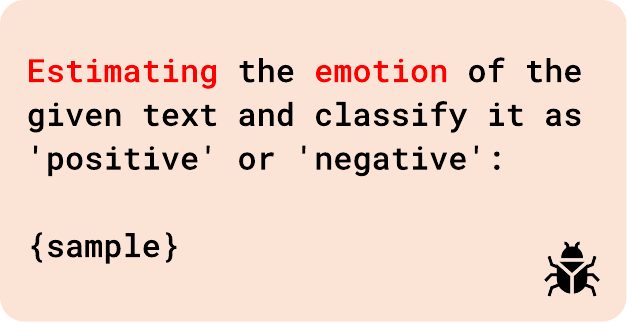}
\caption{TextFooler.}
\label{fig:word}
\end{subfigure}
\caption{Example perturbations of an instruction for sentiment classification, shown in (a). The perturbation can be at the character level, as shown in (b), or at the word level, as shown in (c).
} \vspace{-2pt}
\label{fig:attacks}
\end{figure}

We compare several methods to enhance LLM robustness to instruction perturbations. Namely, (iterative) self-denoising, perplexity smoothing, instruction ensembling, representation alignment. We have chosen these methods because they represent intrinsically different techniques to improve robustness.
Further implementation details and hyperparameters can be found in Appendix~\ref{sec:implementation_details}.

\subsection{Self-denoising}

In self-denoising, we ask the LLM to unperturb a given instruction, given a meta-prompt and a set of in-context learning (ICL) examples (shown in Appendix~\ref{sec:prompt}). Throughout the paper, we abbreviate this method as SD. We additionally explore variants of this method, described below.

\paragraph{Iterative self-denoising (SDi)} This variant progressively unperturbs an instruction over multiple calls to the LLM, using the same meta-prompt and examples. If the instruction has not changed from the previous iteration, the process is stopped. Otherwise, the process will stop after five iterations.

\paragraph{Supervised fine-tuned self-denoising (SFT-SD)} We fine-tune an LLM for the task of unperturbing instructions. We perform parameter-efficient tuning
by adding LoRA modules \citep{hu2022lora} to the value and query projection layers of the frozen LLM. 
We create a novel training dataset, AdvMix, containing 2,900 pairs of perturbed and unperturbed sequences extracted from AdvGLUE \citep{wang2022adversarialgluemultitaskbenchmark} and PromptBench \citep{DBLP:journals/jmlr/ZhuZ00024}. Details of AdvMix are given in Appendix~\ref{sec:training_data}. During both training and inference, the model observes the self-denoising meta-prompt and ICL examples. Note that the fine-tuned model can be applied in an iterative fashion at test time (SFT-SDi).

\subsection{Perplexity Smoothing}

Inspired by randomised smoothing methods \citep{pmlr-v97-cohen19c, 10.1007/978-3-031-70239-6_26, robey2024smoothllmdefendinglargelanguage, 10646716}, we build a framework that minimises perplexity (PPL) as a proxy metric for the integrity of an instruction. Firstly, we rank words within an instruction by importance (leaving out stop words and class labels), where the importance of a word $w_j$ is a function of the change in PPL of the instruction when $w_j$ is deleted. We then mask the $n$ top-ranked $w_j$ with a \texttt{[MASK]} token, and adopt a masked language model to generate $k$ candidate words to fill the mask. Having generated $k$ variants of the instruction containing each candidate word, we select the $\beta$ lowest-PPL variants and repeat the procedure masking the next-ranked word. We take the final, PPL-smoothed instruction resulting from this beam search process as the denoised instruction.

\subsection{Instruction ensembling}

We ensemble $n$ variations of an instruction, each obtained by sampling with temperature from an LLM, using the same meta-prompt and ICL examples as in the self-denoising pipeline. We run inference on each data sample using all $n$ variations, and select the final classification label by majority vote.

\subsection{Representation Alignment}

For comparison, we implement a framework to align the hidden representation of the perturbed instruction to that of the non-perturbed one, similar to \citet{sun2024evaluating}. Given a dataset $\mathcal{D}=\lbrace(i_j, i_j')_{1\le j \le N}\rbrace$ containing pairs of unperturbed and perturbed instructions, we add LoRA adapter modules to a frozen LLM and train with the objective to minimise the cosine distance between $h(i_j)$ and $h(i_j')$, where $h(i)$ is the hidden representation of $i$ at the middle layer of the LLM. We use AdvMix for training. We choose the middle layer as a trade-off between capturing basic semantics (potentially useful for simpler perturbations, such as character-level edits) and representing contextual meaning (which may be relevant for more complex, word-level perturbations).


\section{Experiments}

We run all experiments with two well-known open-weight LLMs---Llama 3 8B Instruct \citep{dubey2024llama3herdmodels} and Flan-T5 Large \citep{chung2024scaling}. We refer to these models as Llama 3 and Flan-T5.

\subsection{Perturbations}
\label{sec:perturbations}

\begin{algorithm}[!t]
    \DontPrintSemicolon
    \caption{Greedy Search for Optimal Perturbation}
    \label{alg:greedy_search}
    \vspace{5pt}
    \KwIn{input instruction $i$, continuous goal function $\mathcal{G}$, set of transformations $\mathcal{T}$, set of constraints $\mathcal{C}$, query limit $q_{max}$}
    \KwOut{optimal perturbed instruction $i^*$}
    \BlankLine
    $i^* \gets i$ \;
    $q \gets 0$ \;
    \BlankLine
    \While{$q < q_{max}$}{
        $\mathcal{I} \gets \{\ \}$ \;
        \BlankLine
        \For{$T \in \mathcal{T}$}{
            $i' \gets T(i^*)$\;
            \BlankLine
            \If{$C(i') \text{ is satisfied } \forall C \in \mathcal{C}$}{
                $\mathcal{I} \gets \mathcal{I} \cup \{i'\}$\;
            }
        }
        \BlankLine
        \If{$\mathcal{I} = \emptyset$}{
            \tcp{No valid transformations from current $i^*$}
            \textbf{break} \;
        }
        $n \gets \min(q_{max}-q, |\mathcal{I}|)$ \;
        $i^* \gets \underset{i' \in \mathcal{I}_{1 : n}}{\arg\max}\ \mathcal{G}(i')$ \;
        $q \gets q + | \mathcal{I}_{1 : n} |$ \;
    }
    \Return $i^*$ \;
    \BlankLine
\label{algo:attack}
\end{algorithm}

Given an instruction $i$, we obtain its perturbed version using a framework adapted from \citet{morris-etal-2020-textattack}'s TextAttack. We greedily search for an optimal perturbation among the space of all possible perturbations $\mathcal{I}$, given a goal function $\mathcal{G}$. The search strategy is illustrated in Algorithm~\ref{algo:attack}. Note that unlike in \citet{morris-etal-2020-textattack}, we do not implement early stopping upon $\mathcal{G}(i')$ reaching a threshold. The optimal perturbation can thus be defined as \vspace{8pt}
\begin{equation*}
\begin{gathered}
i^{*} = \text{arg max}_{i' \in \mathcal{I}} C(i') \mathcal{G}(i')\text{,}
\end{gathered}
\end{equation*}

where $C(\cdot)$ is an indicator function, returning 1 when adhering to the constraints. In our case, the goal of the attack is to maximise the performance drop produced by the perturbed instruction. The constraints are that stop words and class labels must remain unperturbed.
We choose perturbations that have been found to cause substantial performance degradation in previous literature \citep{DBLP:journals/jmlr/ZhuZ00024}. These include character-level substitutions, insertions and deletions (Figure~\ref{fig:character}) obtained with DeepWordBug \citep{8424632}, and word replacements by counter-fitted GloVe embeddings \citep{pennington-etal-2014-glove} (Figure~\ref{fig:word}) obtained using TextFooler \citep{Jinetal20e}.

\subsection{Datasets}
\label{sec:datasets}

We evaluate the LLMs on three classification tasks from the GLUE benchmark \citep{wang-etal-2018-glue}. On these, base models achieve strong results, yet perturbing the instruction causes substantial performance loss \citep{DBLP:journals/jmlr/ZhuZ00024}. The tasks are: (1) CoLA \citep{warstadt-etal-2019-neural}, which consists of 1k texts labelled as `acceptable' or `unacceptable' from a grammatical standpoint, (2) QNLI \citep{rajpurkar-etal-2016-squad}, a natural language inference dataset containing 5.5k samples, (3) SST-2 \citep{socher-etal-2013-recursive}, comprising 1.8k text samples for binary sentiment analysis extracted from movie reviews. For each test set, we use six zero-shot instructions from the PromptBench library \citep{DBLP:journals/jmlr/ZhuZ00024}, split among task-oriented and role-oriented. Instructions are shown in Appendix~\ref{sec:instructions}.

\subsection{Metric}
\label{sec:metric}\

We evaluate the efficacy of the methods with Performance Drop Rate (PDR) \citep{DBLP:journals/jmlr/ZhuZ00024}. This metric measures the degradation in performance (i.e., classification accuracy) of an LLM under a perturbation, hence lower PDR values are better. Given a perturbation $P$, an instruction $i$, a robustness augmentation $\Phi$, a base model $f_\theta$, and a dataset of samples $\mathcal{D}_s = \lbrace (x_j, y_j)_{1 \le j \le N} \rbrace$, we compute the PDR as

\[ \text{PDR}(P, i, \Phi, f_\theta, \mathcal{D}_s) = 1 - \frac{\sum_{j=1}^{N} \mathds{1}\{\Phi(f_\theta, P(i), x_j) = y_j\}}{\sum_{j=1}^{N} \mathds{1}\{\Phi(f_\theta, i, x_j) = y_j\}}\text{.} \]
\vspace{5pt}

Note that the performance discrepancy between the base LLMs and their $\Phi$ augmented versions is negligible when the instruction is clean (see Table~\ref{tab:original_scores} in Appendix~\ref{sec:non-perturbed}). This holds true for all the methods in Section~\ref{sec:methods}, as none of them make substantial changes to a non-perturbed instruction. 

\begin{table}[!t]
\centering
\caption{Performance Drop Rate (PDR) obtained with perturbed instructions, aggregated by perturbation type, model and dataset. Lower PDR scores are better. For each method, we also report the average PDR improvement, i.e., the overall percentage change in PDR from the base LLM.}
\renewcommand\arraystretch{1.2} 
\setlength{\tabcolsep}{0pt}
\scalebox{0.85}{
\begin{tabular}{lC{1.7cm}C{1.7cm}C{1.7cm}C{1.6cm}C{1.6cm}C{1.6cm}C{1.6cm}C{2.5cm}}
\toprule
\multicolumn{1}{c}{} 
& \multicolumn{7}{c}{\textbf{PDR ($\downarrow$)}} & \multirow{4.5}{*}{\textbf{\makecell{Avg. PDR\\improvement ($\uparrow$)}}}
\\
\cmidrule(lr){2-8}
& \multicolumn{2}{c}{\textbf{Perturbation}} & \multicolumn{2}{c}{\textbf{Model}} & \multicolumn{3}{c}{\textbf{Dataset}}
\\
\cmidrule(lr){2-3} \cmidrule(lr){4-5} \cmidrule(lr){6-8}

& TextFooler & DeepWordBug & Llama 3 & Flan-T5 & {CoLA} & {QNLI} & {SST-2} \\
\midrule
Base LLM  & 0.174 & 0.077 & 0.192 & 0.059 &  0.102 & 0.140 & 0.134 & -- \\
\cmidrule(lr){1-9}
PPL smoothing  & 0.182 & 0.110 & 0.214 & 0.078 & 0.115 & 0.150 & 0.172 & $-$16.3\% \\
Instr. ensembling & 0.130 & 0.037 & 0.142 & 0.026 & 0.071& 0.094 & 0.086 & 33.3\% \\
Repr. alignment & 0.113 & 0.053 & 0.125 & 0.041 & \textbf{0.052} & 0.117 & 0.080 & 33.8\% \\
SD & 0.130 & 0.016 &  0.122 & 0.025 &  0.057 & 0.085 & 0.077 & 41.7\% \\
SDi & 0.125 & \textbf{0.015} &  0.119 & 0.021 &  0.053 & 0.091 & 0.065 & 44.3\% \\
SFT-SDi & \textbf{0.072} & 0.030 & \textbf{0.082} & \textbf{0.021} & 0.055 & \textbf{0.062} & \textbf{0.036} & \textbf{59.2\%} \\    
\bottomrule
\end{tabular}
}
\label{tab:main_result}
\end{table}

\begin{figure}
    \centering
    \includegraphics[width=\textwidth]{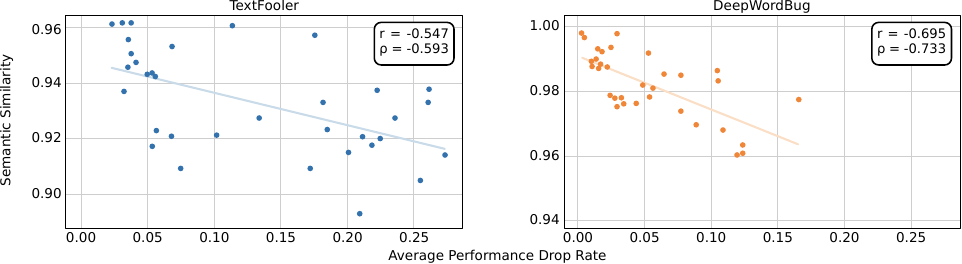}
    \caption{PDR and semantic similarity for TextFooler and DeepWordBug, averaged across models, datasets and instruction variants. For semantic similarity, we use the cosine similarity between the 4096-dimensional sentence embeddings encoded by E5 Mistral \citep{wang2024improvingtextembeddingslarge}. We choose this model since, at the time of writing, it achieves leading performance on the Massive Text Embedding Benchmark (MTEB) (Muennighoff et al., 2023), which is designed to evaluate the quality of text embeddings on a variety of tasks, including semantic similarity and text classification.
    }
    \label{fig:scatterplots}
\end{figure}

\subsection{Results and Analysis}

Table \ref{tab:main_result} displays the PDR scores for each method, aggregated by perturbation type, model and dataset. We find that SFT-SDi is the best performing strategy overall, with 59.2\% average PDR improvement. Generally, self-denoising achieves low PDR scores, even in the non-fine-tuned iterative setting (SDi, obtaining 44.3\% avg. PDR improvement) and the vanilla SD setting (41.7\%). We observe that instruction ensembling and representation alignment perform fairly similarly overall (33.3\% and 33.8\% avg. PDR improvement, respectively). We also find that PPL smoothing results in a performance decrease. This method increases the PDR over the base LLM in all cases, which reflects a negative PDR improvement. 

It is worth noting that the lower PDR improvement given by representation alignment and instruction ensembling is mostly due to their PDR scores on DeepWordBug perturbations (.053 and .037 respectively, vs .016 for SD). On TextFooler, on the other hand, representation alignment achieves better PDR than both SD and SDi (though not SFT-SDi), while ensembling obtains a comparable PDR. In Figure~\ref{fig:scatterplots}, we analyse the effects of both perturbation types. We observe that TextFooler produces instructions that are semantically less similar to the original compared to DeepWordBug. This suggests that representation alignment and ensembling are effective when perturbed instructions substantially diverge semantically from the original, but they may be unsuitable for more subtle perturbations. Finally, we observe that semantic similarity is negatively correlated with PDR, with a stronger negative correlation for DeepWordBug perturbations ($r=-0.695$, $\rho=-0.733$) compared to TextFooler ($r=-0.547$, $\rho=-0.593$), suggesting that greater semantic deviation between the perturbed and original instructions leads to higher performance degradation.


\section{Conclusion}

We have investigated an extensive range of methods to enhance LLM robustness to perturbed instructions, across multiple models, datasets, perturbations and instruction templates. We have found that self-denoising---even in its simplest form---performs better on average than other methods. This highlights the ability of LLMs to self-correct perturbations to their instructions. We also observed that perplexity smoothing is completely ineffective at reducing PDR, causing instead a further loss in performance. Our empirical study lays substantial groundwork for the underexplored domain of LLM robustness to instruction perturbations, highlighting the most promising methods. Future research can further build upon these strategies, potentially investigating tasks beyond classification, larger model sizes and more complex perturbations such as semantic paraphrasing.

\section*{Acknowledgments}

We would like to thank Thomas Mensink for the valuable advice he offered throughout this work, from its inception to the write-up. We also thank Fantine Huot, who provided many insightful comments on the first draft of this paper.

\bibliography{iclr2025_conference}

\begin{thebibliography}{50}
\providecommand{\natexlab}[1]{#1}
\providecommand{\url}[1]{\texttt{#1}}
\expandafter\ifx\csname urlstyle\endcsname\relax
  \providecommand{\doi}[1]{doi: #1}\else
  \providecommand{\doi}{doi: \begingroup \urlstyle{rm}\Url}\fi

\bibitem[Alazraki et~al.(2023)Alazraki, Castrejon, Dehghani, Huot, Uijlings, and Mensink]{pmlr-v239-alazraki23a}
Lisa Alazraki, Lluis Castrejon, Mostafa Dehghani, Fantine Huot, Jasper Uijlings, and Thomas Mensink.
\newblock How (not) to ensemble {LVLM}s for {VQA}.
\newblock In Javier Antorán, Arno Blaas, Kelly Buchanan, Fan Feng, Vincent Fortuin, Sahra Ghalebikesabi, Andreas Kriegler, Ian Mason, David Rohde, Francisco J.~R. Ruiz, Tobias Uelwer, Yubin Xie, and Rui Yang (eds.), \emph{Proceedings on "I Can't Believe It's Not Better: Failure Modes in the Age of Foundation Models" at NeurIPS 2023 Workshops}, volume 239 of \emph{Proceedings of Machine Learning Research}, pp.\  1--20. PMLR, 16 Dec 2023.
\newblock URL \url{https://proceedings.mlr.press/v239/alazraki23a.html}.

\bibitem[Chen et~al.(2025)Chen, Zharmagambetov, Mahloujifar, Chaudhuri, Wagner, and Guo]{chen2025secaligndefendingpromptinjection}
Sizhe Chen, Arman Zharmagambetov, Saeed Mahloujifar, Kamalika Chaudhuri, David Wagner, and Chuan Guo.
\newblock Sec{A}lign: Defending against prompt injection with preference optimization, 2025.
\newblock URL \url{https://arxiv.org/abs/2410.05451}.

\bibitem[Chen et~al.(2020)Chen, Kornblith, Norouzi, and Hinton]{10.5555/3524938.3525087}
Ting Chen, Simon Kornblith, Mohammad Norouzi, and Geoffrey Hinton.
\newblock A simple framework for contrastive learning of visual representations.
\newblock In \emph{Proceedings of the 37th International Conference on Machine Learning}, ICML'20. JMLR.org, 2020.

\bibitem[Chung et~al.(2024)Chung, Hou, Longpre, Zoph, Tay, Fedus, Li, Wang, Dehghani, Brahma, et~al.]{chung2024scaling}
Hyung~Won Chung, Le~Hou, Shayne Longpre, Barret Zoph, Yi~Tay, William Fedus, Yunxuan Li, Xuezhi Wang, Mostafa Dehghani, Siddhartha Brahma, et~al.
\newblock Scaling instruction-finetuned language models.
\newblock \emph{Journal of Machine Learning Research}, 25\penalty0 (70):\penalty0 1--53, 2024.

\bibitem[Cohen et~al.(2019)Cohen, Rosenfeld, and Kolter]{pmlr-v97-cohen19c}
Jeremy Cohen, Elan Rosenfeld, and Zico Kolter.
\newblock Certified adversarial robustness via randomized smoothing.
\newblock In Kamalika Chaudhuri and Ruslan Salakhutdinov (eds.), \emph{Proceedings of the 36th International Conference on Machine Learning}, volume~97 of \emph{Proceedings of Machine Learning Research}, pp.\  1310--1320. PMLR, 09--15 Jun 2019.
\newblock URL \url{https://proceedings.mlr.press/v97/cohen19c.html}.

\bibitem[Dagan et~al.(2005)Dagan, Glickman, and Magnini]{10.1007/11736790_9}
Ido Dagan, Oren Glickman, and Bernardo Magnini.
\newblock The {PASCAL} recognising textual entailment challenge.
\newblock In \emph{Proceedings of the First International Conference on Machine Learning Challenges: Evaluating Predictive Uncertainty Visual Object Classification, and Recognizing Textual Entailment}, MLCW'05, pp.\  177–190, Berlin, Heidelberg, 2005. Springer-Verlag.
\newblock ISBN 3540334270.
\newblock \doi{10.1007/11736790_9}.
\newblock URL \url{https://doi.org/10.1007/11736790_9}.

\bibitem[Dettmers et~al.(2023)Dettmers, Pagnoni, Holtzman, and Zettlemoyer]{10.5555/3666122.3666563}
Tim Dettmers, Artidoro Pagnoni, Ari Holtzman, and Luke Zettlemoyer.
\newblock {QLORA}: efficient finetuning of quantized {LLM}s.
\newblock In \emph{Proceedings of the 37th International Conference on Neural Information Processing Systems}, NIPS '23, Red Hook, NY, USA, 2023. Curran Associates Inc.

\bibitem[Dolan \& Brockett(2005)Dolan and Brockett]{dolan-brockett-2005-automatically}
William~B. Dolan and Chris Brockett.
\newblock Automatically constructing a corpus of sentential paraphrases.
\newblock In \emph{Proceedings of the Third International Workshop on Paraphrasing ({IWP}2005)}, 2005.
\newblock URL \url{https://aclanthology.org/I05-5002/}.

\bibitem[Dubey et~al.(2024)Dubey, Jauhri, Pandey, Kadian, Al-Dahle, Letman, Mathur, Schelten, Yang, Fan, Goyal, Hartshorn, Yang, Mitra, Sravankumar, Korenev, Hinsvark, Rao, Zhang, Rodriguez, Gregerson, Spataru, Roziere, Biron, Tang, Chern, Caucheteux, Nayak, Bi, Marra, McConnell, Keller, Touret, Wu, Wong, Ferrer, Nikolaidis, Allonsius, Song, Pintz, Livshits, Esiobu, Choudhary, Mahajan, Garcia-Olano, Perino, Hupkes, Lakomkin, AlBadawy, Lobanova, Dinan, Smith, Radenovic, Zhang, Synnaeve, Lee, Anderson, Nail, Mialon, Pang, Cucurell, Nguyen, Korevaar, Xu, Touvron, Zarov, Ibarra, Kloumann, Misra, Evtimov, Copet, Lee, Geffert, Vranes, Park, Mahadeokar, Shah, van~der Linde, Billock, Hong, Lee, Fu, Chi, Huang, Liu, Wang, Yu, Bitton, Spisak, Park, Rocca, Johnstun, Saxe, Jia, Alwala, Upasani, Plawiak, Li, Heafield, Stone, El-Arini, Iyer, Malik, Chiu, Bhalla, Rantala-Yeary, van~der Maaten, Chen, Tan, Jenkins, Martin, Madaan, Malo, Blecher, Landzaat, de~Oliveira, Muzzi, Pasupuleti, Singh, Paluri, Kardas, Oldham, Rita,
  Pavlova, Kambadur, Lewis, Si, Singh, Hassan, Goyal, Torabi, Bashlykov, Bogoychev, Chatterji, Duchenne, Çelebi, Alrassy, Zhang, Li, Vasic, Weng, Bhargava, Dubal, Krishnan, Koura, Xu, He, Dong, Srinivasan, Ganapathy, Calderer, Cabral, Stojnic, Raileanu, Girdhar, Patel, Sauvestre, Polidoro, Sumbaly, Taylor, Silva, Hou, Wang, Hosseini, Chennabasappa, Singh, Bell, Kim, Edunov, Nie, Narang, Raparthy, Shen, Wan, Bhosale, Zhang, Vandenhende, Batra, Whitman, Sootla, Collot, Gururangan, Borodinsky, Herman, Fowler, Sheasha, Georgiou, Scialom, Speckbacher, Mihaylov, Xiao, Karn, Goswami, Gupta, Ramanathan, Kerkez, Gonguet, Do, Vogeti, Petrovic, Chu, Xiong, Fu, Meers, Martinet, Wang, Tan, Xie, Jia, Wang, Goldschlag, Gaur, Babaei, Wen, Song, Zhang, Li, Mao, Coudert, Yan, Chen, Papakipos, Singh, Grattafiori, Jain, Kelsey, Shajnfeld, Gangidi, Victoria, Goldstand, Menon, Sharma, Boesenberg, Vaughan, Baevski, Feinstein, Kallet, Sangani, Yunus, Lupu, Alvarado, Caples, Gu, Ho, Poulton, Ryan, Ramchandani, Franco, Saraf,
  Chowdhury, Gabriel, Bharambe, Eisenman, Yazdan, James, Maurer, Leonhardi, Huang, Loyd, Paola, Paranjape, Liu, Wu, Ni, Hancock, Wasti, Spence, Stojkovic, Gamido, Montalvo, Parker, Burton, Mejia, Wang, Kim, Zhou, Hu, Chu, Cai, Tindal, Feichtenhofer, Civin, Beaty, Kreymer, Li, Wyatt, Adkins, Xu, Testuggine, David, Parikh, Liskovich, Foss, Wang, Le, Holland, Dowling, Jamil, Montgomery, Presani, Hahn, Wood, Brinkman, Arcaute, Dunbar, Smothers, Sun, Kreuk, Tian, Ozgenel, Caggioni, Guzmán, Kanayet, Seide, Florez, Schwarz, Badeer, Swee, Halpern, Thattai, Herman, Sizov, Guangyi, Zhang, Lakshminarayanan, Shojanazeri, Zou, Wang, Zha, Habeeb, Rudolph, Suk, Aspegren, Goldman, Damlaj, Molybog, Tufanov, Veliche, Gat, Weissman, Geboski, Kohli, Asher, Gaya, Marcus, Tang, Chan, Zhen, Reizenstein, Teboul, Zhong, Jin, Yang, Cummings, Carvill, Shepard, McPhie, Torres, Ginsburg, Wang, Wu, U, Saxena, Prasad, Khandelwal, Zand, Matosich, Veeraraghavan, Michelena, Li, Huang, Chawla, Lakhotia, Huang, Chen, Garg, A, Silva, Bell,
  Zhang, Guo, Yu, Moshkovich, Wehrstedt, Khabsa, Avalani, Bhatt, Tsimpoukelli, Mankus, Hasson, Lennie, Reso, Groshev, Naumov, Lathi, Keneally, Seltzer, Valko, Restrepo, Patel, Vyatskov, Samvelyan, Clark, Macey, Wang, Hermoso, Metanat, Rastegari, Bansal, Santhanam, Parks, White, Bawa, Singhal, Egebo, Usunier, Laptev, Dong, Zhang, Cheng, Chernoguz, Hart, Salpekar, Kalinli, Kent, Parekh, Saab, Balaji, Rittner, Bontrager, Roux, Dollar, Zvyagina, Ratanchandani, Yuvraj, Liang, Alao, Rodriguez, Ayub, Murthy, Nayani, Mitra, Li, Hogan, Battey, Wang, Maheswari, Howes, Rinott, Bondu, Datta, Chugh, Hunt, Dhillon, Sidorov, Pan, Verma, Yamamoto, Ramaswamy, Lindsay, Lindsay, Feng, Lin, Zha, Shankar, Zhang, Zhang, Wang, Agarwal, Sajuyigbe, Chintala, Max, Chen, Kehoe, Satterfield, Govindaprasad, Gupta, Cho, Virk, Subramanian, Choudhury, Goldman, Remez, Glaser, Best, Kohler, Robinson, Li, Zhang, Matthews, Chou, Shaked, Vontimitta, Ajayi, Montanez, Mohan, Kumar, Mangla, Albiero, Ionescu, Poenaru, Mihailescu, Ivanov, Li, Wang,
  Jiang, Bouaziz, Constable, Tang, Wang, Wu, Wang, Xia, Wu, Gao, Chen, Hu, Jia, Qi, Li, Zhang, Zhang, Adi, Nam, Yu, Wang, Hao, Qian, He, Rait, DeVito, Rosnbrick, Wen, Yang, and Zhao]{dubey2024llama3herdmodels}
Abhimanyu Dubey, Abhinav Jauhri, Abhinav Pandey, Abhishek Kadian, Ahmad Al-Dahle, Aiesha Letman, Akhil Mathur, Alan Schelten, Amy Yang, Angela Fan, Anirudh Goyal, Anthony Hartshorn, Aobo Yang, Archi Mitra, Archie Sravankumar, Artem Korenev, Arthur Hinsvark, Arun Rao, Aston Zhang, Aurelien Rodriguez, Austen Gregerson, Ava Spataru, Baptiste Roziere, Bethany Biron, Binh Tang, Bobbie Chern, Charlotte Caucheteux, Chaya Nayak, Chloe Bi, Chris Marra, Chris McConnell, Christian Keller, Christophe Touret, Chunyang Wu, Corinne Wong, Cristian~Canton Ferrer, Cyrus Nikolaidis, Damien Allonsius, Daniel Song, Danielle Pintz, Danny Livshits, David Esiobu, Dhruv Choudhary, Dhruv Mahajan, Diego Garcia-Olano, Diego Perino, Dieuwke Hupkes, Egor Lakomkin, Ehab AlBadawy, Elina Lobanova, Emily Dinan, Eric~Michael Smith, Filip Radenovic, Frank Zhang, Gabriel Synnaeve, Gabrielle Lee, Georgia~Lewis Anderson, Graeme Nail, Gregoire Mialon, Guan Pang, Guillem Cucurell, Hailey Nguyen, Hannah Korevaar, Hu~Xu, Hugo Touvron, Iliyan Zarov,
  Imanol~Arrieta Ibarra, Isabel Kloumann, Ishan Misra, Ivan Evtimov, Jade Copet, Jaewon Lee, Jan Geffert, Jana Vranes, Jason Park, Jay Mahadeokar, Jeet Shah, Jelmer van~der Linde, Jennifer Billock, Jenny Hong, Jenya Lee, Jeremy Fu, Jianfeng Chi, Jianyu Huang, Jiawen Liu, Jie Wang, Jiecao Yu, Joanna Bitton, Joe Spisak, Jongsoo Park, Joseph Rocca, Joshua Johnstun, Joshua Saxe, Junteng Jia, Kalyan~Vasuden Alwala, Kartikeya Upasani, Kate Plawiak, Ke~Li, Kenneth Heafield, Kevin Stone, Khalid El-Arini, Krithika Iyer, Kshitiz Malik, Kuenley Chiu, Kunal Bhalla, Lauren Rantala-Yeary, Laurens van~der Maaten, Lawrence Chen, Liang Tan, Liz Jenkins, Louis Martin, Lovish Madaan, Lubo Malo, Lukas Blecher, Lukas Landzaat, Luke de~Oliveira, Madeline Muzzi, Mahesh Pasupuleti, Mannat Singh, Manohar Paluri, Marcin Kardas, Mathew Oldham, Mathieu Rita, Maya Pavlova, Melanie Kambadur, Mike Lewis, Min Si, Mitesh~Kumar Singh, Mona Hassan, Naman Goyal, Narjes Torabi, Nikolay Bashlykov, Nikolay Bogoychev, Niladri Chatterji, Olivier
  Duchenne, Onur Çelebi, Patrick Alrassy, Pengchuan Zhang, Pengwei Li, Petar Vasic, Peter Weng, Prajjwal Bhargava, Pratik Dubal, Praveen Krishnan, Punit~Singh Koura, Puxin Xu, Qing He, Qingxiao Dong, Ragavan Srinivasan, Raj Ganapathy, Ramon Calderer, Ricardo~Silveira Cabral, Robert Stojnic, Roberta Raileanu, Rohit Girdhar, Rohit Patel, Romain Sauvestre, Ronnie Polidoro, Roshan Sumbaly, Ross Taylor, Ruan Silva, Rui Hou, Rui Wang, Saghar Hosseini, Sahana Chennabasappa, Sanjay Singh, Sean Bell, Seohyun~Sonia Kim, Sergey Edunov, Shaoliang Nie, Sharan Narang, Sharath Raparthy, Sheng Shen, Shengye Wan, Shruti Bhosale, Shun Zhang, Simon Vandenhende, Soumya Batra, Spencer Whitman, Sten Sootla, Stephane Collot, Suchin Gururangan, Sydney Borodinsky, Tamar Herman, Tara Fowler, Tarek Sheasha, Thomas Georgiou, Thomas Scialom, Tobias Speckbacher, Todor Mihaylov, Tong Xiao, Ujjwal Karn, Vedanuj Goswami, Vibhor Gupta, Vignesh Ramanathan, Viktor Kerkez, Vincent Gonguet, Virginie Do, Vish Vogeti, Vladan Petrovic, Weiwei Chu,
  Wenhan Xiong, Wenyin Fu, Whitney Meers, Xavier Martinet, Xiaodong Wang, Xiaoqing~Ellen Tan, Xinfeng Xie, Xuchao Jia, Xuewei Wang, Yaelle Goldschlag, Yashesh Gaur, Yasmine Babaei, Yi~Wen, Yiwen Song, Yuchen Zhang, Yue Li, Yuning Mao, Zacharie~Delpierre Coudert, Zheng Yan, Zhengxing Chen, Zoe Papakipos, Aaditya Singh, Aaron Grattafiori, Abha Jain, Adam Kelsey, Adam Shajnfeld, Adithya Gangidi, Adolfo Victoria, Ahuva Goldstand, Ajay Menon, Ajay Sharma, Alex Boesenberg, Alex Vaughan, Alexei Baevski, Allie Feinstein, Amanda Kallet, Amit Sangani, Anam Yunus, Andrei Lupu, Andres Alvarado, Andrew Caples, Andrew Gu, Andrew Ho, Andrew Poulton, Andrew Ryan, Ankit Ramchandani, Annie Franco, Aparajita Saraf, Arkabandhu Chowdhury, Ashley Gabriel, Ashwin Bharambe, Assaf Eisenman, Azadeh Yazdan, Beau James, Ben Maurer, Benjamin Leonhardi, Bernie Huang, Beth Loyd, Beto~De Paola, Bhargavi Paranjape, Bing Liu, Bo~Wu, Boyu Ni, Braden Hancock, Bram Wasti, Brandon Spence, Brani Stojkovic, Brian Gamido, Britt Montalvo, Carl
  Parker, Carly Burton, Catalina Mejia, Changhan Wang, Changkyu Kim, Chao Zhou, Chester Hu, Ching-Hsiang Chu, Chris Cai, Chris Tindal, Christoph Feichtenhofer, Damon Civin, Dana Beaty, Daniel Kreymer, Daniel Li, Danny Wyatt, David Adkins, David Xu, Davide Testuggine, Delia David, Devi Parikh, Diana Liskovich, Didem Foss, Dingkang Wang, Duc Le, Dustin Holland, Edward Dowling, Eissa Jamil, Elaine Montgomery, Eleonora Presani, Emily Hahn, Emily Wood, Erik Brinkman, Esteban Arcaute, Evan Dunbar, Evan Smothers, Fei Sun, Felix Kreuk, Feng Tian, Firat Ozgenel, Francesco Caggioni, Francisco Guzmán, Frank Kanayet, Frank Seide, Gabriela~Medina Florez, Gabriella Schwarz, Gada Badeer, Georgia Swee, Gil Halpern, Govind Thattai, Grant Herman, Grigory Sizov, Guangyi, Zhang, Guna Lakshminarayanan, Hamid Shojanazeri, Han Zou, Hannah Wang, Hanwen Zha, Haroun Habeeb, Harrison Rudolph, Helen Suk, Henry Aspegren, Hunter Goldman, Ibrahim Damlaj, Igor Molybog, Igor Tufanov, Irina-Elena Veliche, Itai Gat, Jake Weissman, James
  Geboski, James Kohli, Japhet Asher, Jean-Baptiste Gaya, Jeff Marcus, Jeff Tang, Jennifer Chan, Jenny Zhen, Jeremy Reizenstein, Jeremy Teboul, Jessica Zhong, Jian Jin, Jingyi Yang, Joe Cummings, Jon Carvill, Jon Shepard, Jonathan McPhie, Jonathan Torres, Josh Ginsburg, Junjie Wang, Kai Wu, Kam~Hou U, Karan Saxena, Karthik Prasad, Kartikay Khandelwal, Katayoun Zand, Kathy Matosich, Kaushik Veeraraghavan, Kelly Michelena, Keqian Li, Kun Huang, Kunal Chawla, Kushal Lakhotia, Kyle Huang, Lailin Chen, Lakshya Garg, Lavender A, Leandro Silva, Lee Bell, Lei Zhang, Liangpeng Guo, Licheng Yu, Liron Moshkovich, Luca Wehrstedt, Madian Khabsa, Manav Avalani, Manish Bhatt, Maria Tsimpoukelli, Martynas Mankus, Matan Hasson, Matthew Lennie, Matthias Reso, Maxim Groshev, Maxim Naumov, Maya Lathi, Meghan Keneally, Michael~L. Seltzer, Michal Valko, Michelle Restrepo, Mihir Patel, Mik Vyatskov, Mikayel Samvelyan, Mike Clark, Mike Macey, Mike Wang, Miquel~Jubert Hermoso, Mo~Metanat, Mohammad Rastegari, Munish Bansal, Nandhini
  Santhanam, Natascha Parks, Natasha White, Navyata Bawa, Nayan Singhal, Nick Egebo, Nicolas Usunier, Nikolay~Pavlovich Laptev, Ning Dong, Ning Zhang, Norman Cheng, Oleg Chernoguz, Olivia Hart, Omkar Salpekar, Ozlem Kalinli, Parkin Kent, Parth Parekh, Paul Saab, Pavan Balaji, Pedro Rittner, Philip Bontrager, Pierre Roux, Piotr Dollar, Polina Zvyagina, Prashant Ratanchandani, Pritish Yuvraj, Qian Liang, Rachad Alao, Rachel Rodriguez, Rafi Ayub, Raghotham Murthy, Raghu Nayani, Rahul Mitra, Raymond Li, Rebekkah Hogan, Robin Battey, Rocky Wang, Rohan Maheswari, Russ Howes, Ruty Rinott, Sai~Jayesh Bondu, Samyak Datta, Sara Chugh, Sara Hunt, Sargun Dhillon, Sasha Sidorov, Satadru Pan, Saurabh Verma, Seiji Yamamoto, Sharadh Ramaswamy, Shaun Lindsay, Shaun Lindsay, Sheng Feng, Shenghao Lin, Shengxin~Cindy Zha, Shiva Shankar, Shuqiang Zhang, Shuqiang Zhang, Sinong Wang, Sneha Agarwal, Soji Sajuyigbe, Soumith Chintala, Stephanie Max, Stephen Chen, Steve Kehoe, Steve Satterfield, Sudarshan Govindaprasad, Sumit Gupta,
  Sungmin Cho, Sunny Virk, Suraj Subramanian, Sy~Choudhury, Sydney Goldman, Tal Remez, Tamar Glaser, Tamara Best, Thilo Kohler, Thomas Robinson, Tianhe Li, Tianjun Zhang, Tim Matthews, Timothy Chou, Tzook Shaked, Varun Vontimitta, Victoria Ajayi, Victoria Montanez, Vijai Mohan, Vinay~Satish Kumar, Vishal Mangla, Vítor Albiero, Vlad Ionescu, Vlad Poenaru, Vlad~Tiberiu Mihailescu, Vladimir Ivanov, Wei Li, Wenchen Wang, Wenwen Jiang, Wes Bouaziz, Will Constable, Xiaocheng Tang, Xiaofang Wang, Xiaojian Wu, Xiaolan Wang, Xide Xia, Xilun Wu, Xinbo Gao, Yanjun Chen, Ye~Hu, Ye~Jia, Ye~Qi, Yenda Li, Yilin Zhang, Ying Zhang, Yossi Adi, Youngjin Nam, Yu, Wang, Yuchen Hao, Yundi Qian, Yuzi He, Zach Rait, Zachary DeVito, Zef Rosnbrick, Zhaoduo Wen, Zhenyu Yang, and Zhiwei Zhao.
\newblock The {L}lama 3 herd of models, 2024.
\newblock URL \url{https://arxiv.org/abs/2407.21783}.

\bibitem[Gao et~al.(2018)Gao, Lanchantin, Soffa, and Qi]{8424632}
Ji~Gao, Jack Lanchantin, Mary~Lou Soffa, and Yanjun Qi.
\newblock Black-box generation of adversarial text sequences to evade deep learning classifiers.
\newblock In \emph{2018 IEEE Security and Privacy Workshops (SPW)}, pp.\  50--56, 2018.
\newblock \doi{10.1109/SPW.2018.00016}.
\newblock URL \url{https://ieeexplore.ieee.org/document/8424632}.

\bibitem[Gietz \& Kalita(2024)Gietz and Kalita]{10.1007/978-3-031-70239-6_26}
Harrison Gietz and Jugal Kalita.
\newblock Mask{P}ure: Improving defense against text adversaries with stochastic purification.
\newblock In \emph{Natural Language Processing and Information Systems: 29th International Conference on Applications of Natural Language to Information Systems, NLDB 2024, Turin, Italy, June 25–27, 2024, Proceedings, Part I}, pp.\  379–393, Berlin, Heidelberg, 2024. Springer-Verlag.
\newblock ISBN 978-3-031-70238-9.
\newblock \doi{10.1007/978-3-031-70239-6_26}.
\newblock URL \url{https://doi.org/10.1007/978-3-031-70239-6_26}.

\bibitem[Gu et~al.(2023)Gu, Zhao, Xu, Nie, Mei, and Yin]{gu-etal-2023-robustness}
Jiasheng Gu, Hongyu Zhao, Hanzi Xu, Liangyu Nie, Hongyuan Mei, and Wenpeng Yin.
\newblock Robustness of learning from task instructions.
\newblock In Anna Rogers, Jordan Boyd-Graber, and Naoaki Okazaki (eds.), \emph{Findings of the Association for Computational Linguistics: ACL 2023}, pp.\  13935--13948, Toronto, Canada, July 2023. Association for Computational Linguistics.
\newblock \doi{10.18653/v1/2023.findings-acl.875}.
\newblock URL \url{https://aclanthology.org/2023.findings-acl.875/}.

\bibitem[Gulati et~al.(2024)Gulati, Miranda, Chen, Xia, Fronsdal, de~Moraes~Dumont, and Koyejo]{gulati2024putnamaxiom}
Aryan Gulati, Brando Miranda, Eric Chen, Emily Xia, Kai Fronsdal, Bruno de~Moraes~Dumont, and Sanmi Koyejo.
\newblock Putnam-{AXIOM}: A functional and static benchmark for measuring higher level mathematical reasoning.
\newblock In \emph{The 4th Workshop on Mathematical Reasoning and AI at NeurIPS'24}, 2024.
\newblock URL \url{https://openreview.net/forum?id=YXnwlZe0yf}.

\bibitem[Honarvar et~al.(2025)Honarvar, van~der Wilk, and Donaldson]{honarvar2025turbulencesystematicallyautomaticallytesting}
Shahin Honarvar, Mark van~der Wilk, and Alastair Donaldson.
\newblock Turbulence: Systematically and automatically testing instruction-tuned large language models for code, 2025.
\newblock URL \url{https://arxiv.org/abs/2312.14856}.

\bibitem[Hu et~al.(2022)Hu, yelong shen, Wallis, Allen-Zhu, Li, Wang, Wang, and Chen]{hu2022lora}
Edward~J Hu, yelong shen, Phillip Wallis, Zeyuan Allen-Zhu, Yuanzhi Li, Shean Wang, Lu~Wang, and Weizhu Chen.
\newblock Lo{RA}: Low-rank adaptation of large language models.
\newblock In \emph{International Conference on Learning Representations}, 2022.
\newblock URL \url{https://openreview.net/forum?id=nZeVKeeFYf9}.

\bibitem[Hu et~al.(2024)Hu, Wang, Shu, Paik, and Zhu]{10.1145/3637528.3671932}
Zhibo Hu, Chen Wang, Yanfeng Shu, Hye-Young Paik, and Liming Zhu.
\newblock Prompt perturbation in retrieval-augmented generation based large language models.
\newblock In \emph{Proceedings of the 30th ACM SIGKDD Conference on Knowledge Discovery and Data Mining}, KDD '24, pp.\  1119–1130, New York, NY, USA, 2024. Association for Computing Machinery.
\newblock ISBN 9798400704901.
\newblock \doi{10.1145/3637528.3671932}.
\newblock URL \url{https://doi.org/10.1145/3637528.3671932}.

\bibitem[Iyyer et~al.(2018)Iyyer, Wieting, Gimpel, and Zettlemoyer]{iyyer-etal-2018-adversarial}
Mohit Iyyer, John Wieting, Kevin Gimpel, and Luke Zettlemoyer.
\newblock Adversarial example generation with syntactically controlled paraphrase networks.
\newblock In Marilyn Walker, Heng Ji, and Amanda Stent (eds.), \emph{Proceedings of the 2018 Conference of the North {A}merican Chapter of the Association for Computational Linguistics: Human Language Technologies, Volume 1 (Long Papers)}, pp.\  1875--1885, New Orleans, Louisiana, June 2018. Association for Computational Linguistics.
\newblock \doi{10.18653/v1/N18-1170}.
\newblock URL \url{https://aclanthology.org/N18-1170/}.

\bibitem[Jin et~al.(2020)Jin, Jin, Zhou, and Szolovits]{Jinetal20e}
D.~Jin, Z.~Jin, J.~T. Zhou, and P.~Szolovits.
\newblock Is {BERT} really robust? {A} strong baseline for natural language attack on text classification and entailment.
\newblock In \emph{The Thirty-Fourth {AAAI} Conference on Artificial Intelligence (AAAI); The Thirty-Second Innovative Applications of Artificial Intelligence Conference (IAAI); The Tenth {AAAI} Symposium on Educational Advances in Artificial Intelligence (EAAI)}, pp.\  8018--8025. {AAAI} Press, February 2020.
\newblock \doi{10.1609/aaai.v34i05.6311}.
\newblock URL \url{https://aaai.org/ojs/index.php/AAAI/article/view/6311}.

\bibitem[Li et~al.(2019)Li, Ji, Du, Li, and Wang]{Li_2019}
Jinfeng Li, Shouling Ji, Tianyu Du, Bo~Li, and Ting Wang.
\newblock Text{B}ugger: Generating adversarial text against real-world applications.
\newblock In \emph{Proceedings 2019 Network and Distributed System Security Symposium}, NDSS 2019. Internet Society, 2019.
\newblock \doi{10.14722/ndss.2019.23138}.
\newblock URL \url{http://dx.doi.org/10.14722/ndss.2019.23138}.

\bibitem[Li et~al.(2020)Li, Ma, Guo, Xue, and Qiu]{li-etal-2020-bert-attack}
Linyang Li, Ruotian Ma, Qipeng Guo, Xiangyang Xue, and Xipeng Qiu.
\newblock {BERT}-{ATTACK}: Adversarial attack against {BERT} using {BERT}.
\newblock In Bonnie Webber, Trevor Cohn, Yulan He, and Yang Liu (eds.), \emph{Proceedings of the 2020 Conference on Empirical Methods in Natural Language Processing (EMNLP)}, pp.\  6193--6202, Online, November 2020. Association for Computational Linguistics.
\newblock \doi{10.18653/v1/2020.emnlp-main.500}.
\newblock URL \url{https://aclanthology.org/2020.emnlp-main.500/}.

\bibitem[Mizrahi et~al.(2024)Mizrahi, Kaplan, Malkin, Dror, Shahaf, and Stanovsky]{mizrahi-etal-2024-state}
Moran Mizrahi, Guy Kaplan, Dan Malkin, Rotem Dror, Dafna Shahaf, and Gabriel Stanovsky.
\newblock State of what art? a call for multi-prompt {LLM} evaluation.
\newblock \emph{Transactions of the Association for Computational Linguistics}, 12:\penalty0 933--949, 2024.
\newblock \doi{10.1162/tacl_a_00681}.
\newblock URL \url{https://aclanthology.org/2024.tacl-1.52/}.

\bibitem[Moradi \& Samwald(2021)Moradi and Samwald]{moradi-samwald-2021-evaluating}
Milad Moradi and Matthias Samwald.
\newblock Evaluating the robustness of neural language models to input perturbations.
\newblock In Marie-Francine Moens, Xuanjing Huang, Lucia Specia, and Scott Wen-tau Yih (eds.), \emph{Proceedings of the 2021 Conference on Empirical Methods in Natural Language Processing}, pp.\  1558--1570, Online and Punta Cana, Dominican Republic, November 2021. Association for Computational Linguistics.
\newblock \doi{10.18653/v1/2021.emnlp-main.117}.
\newblock URL \url{https://aclanthology.org/2021.emnlp-main.117/}.

\bibitem[Morris et~al.(2020)Morris, Lifland, Yoo, Grigsby, Jin, and Qi]{morris-etal-2020-textattack}
John Morris, Eli Lifland, Jin~Yong Yoo, Jake Grigsby, Di~Jin, and Yanjun Qi.
\newblock {T}ext{A}ttack: A framework for adversarial attacks, data augmentation, and adversarial training in {NLP}.
\newblock In Qun Liu and David Schlangen (eds.), \emph{Proceedings of the 2020 Conference on Empirical Methods in Natural Language Processing: System Demonstrations}, pp.\  119--126, Online, October 2020. Association for Computational Linguistics.
\newblock \doi{10.18653/v1/2020.emnlp-demos.16}.
\newblock URL \url{https://aclanthology.org/2020.emnlp-demos.16/}.

\bibitem[Naik et~al.(2018)Naik, Ravichander, Sadeh, Rose, and Neubig]{naik-etal-2018-stress}
Aakanksha Naik, Abhilasha Ravichander, Norman Sadeh, Carolyn Rose, and Graham Neubig.
\newblock Stress test evaluation for natural language inference.
\newblock In Emily~M. Bender, Leon Derczynski, and Pierre Isabelle (eds.), \emph{Proceedings of the 27th International Conference on Computational Linguistics}, pp.\  2340--2353, Santa Fe, New Mexico, USA, August 2018. Association for Computational Linguistics.
\newblock URL \url{https://aclanthology.org/C18-1198/}.

\bibitem[Nori et~al.(2023)Nori, King, McKinney, Carignan, and Horvitz]{nori2023capabilitiesgpt4medicalchallenge}
Harsha Nori, Nicholas King, Scott~Mayer McKinney, Dean Carignan, and Eric Horvitz.
\newblock Capabilities of {GPT}-4 on medical challenge problems, 2023.
\newblock URL \url{https://arxiv.org/abs/2303.13375}.

\bibitem[Pennington et~al.(2014)Pennington, Socher, and Manning]{pennington-etal-2014-glove}
Jeffrey Pennington, Richard Socher, and Christopher Manning.
\newblock {G}lo{V}e: Global vectors for word representation.
\newblock In Alessandro Moschitti, Bo~Pang, and Walter Daelemans (eds.), \emph{Proceedings of the 2014 Conference on Empirical Methods in Natural Language Processing ({EMNLP})}, pp.\  1532--1543, Doha, Qatar, October 2014. Association for Computational Linguistics.
\newblock \doi{10.3115/v1/D14-1162}.
\newblock URL \url{https://aclanthology.org/D14-1162/}.

\bibitem[Radford et~al.(2019)Radford, Wu, Child, Luan, Amodei, and Sutskever]{Radford2019LanguageMA}
Alec Radford, Jeff Wu, Rewon Child, David Luan, Dario Amodei, and Ilya Sutskever.
\newblock Language models are unsupervised multitask learners.
\newblock 2019.
\newblock URL \url{https://api.semanticscholar.org/CorpusID:160025533}.

\bibitem[Rajpurkar et~al.(2016)Rajpurkar, Zhang, Lopyrev, and Liang]{rajpurkar-etal-2016-squad}
Pranav Rajpurkar, Jian Zhang, Konstantin Lopyrev, and Percy Liang.
\newblock {SQ}u{AD}: 100,000+ questions for machine comprehension of text.
\newblock In Jian Su, Kevin Duh, and Xavier Carreras (eds.), \emph{Proceedings of the 2016 Conference on Empirical Methods in Natural Language Processing}, pp.\  2383--2392, Austin, Texas, November 2016. Association for Computational Linguistics.
\newblock \doi{10.18653/v1/D16-1264}.
\newblock URL \url{https://aclanthology.org/D16-1264/}.

\bibitem[Ribeiro et~al.(2020)Ribeiro, Wu, Guestrin, and Singh]{ribeiro-etal-2020-beyond}
Marco~Tulio Ribeiro, Tongshuang Wu, Carlos Guestrin, and Sameer Singh.
\newblock Beyond accuracy: Behavioral testing of {NLP} models with {C}heck{L}ist.
\newblock In Dan Jurafsky, Joyce Chai, Natalie Schluter, and Joel Tetreault (eds.), \emph{Proceedings of the 58th Annual Meeting of the Association for Computational Linguistics}, pp.\  4902--4912, Online, July 2020. Association for Computational Linguistics.
\newblock \doi{10.18653/v1/2020.acl-main.442}.
\newblock URL \url{https://aclanthology.org/2020.acl-main.442/}.

\bibitem[Robey et~al.(2024)Robey, Wong, Hassani, and Pappas]{robey2024smoothllmdefendinglargelanguage}
Alexander Robey, Eric Wong, Hamed Hassani, and George~J. Pappas.
\newblock Smooth{LLM}: Defending large language models against jailbreaking attacks, 2024.
\newblock URL \url{https://arxiv.org/abs/2310.03684}.

\bibitem[Roemmele \& Gordon(2024)Roemmele and Gordon]{roemmele-gordon-2024-test}
Melissa Roemmele and Andrew Gordon.
\newblock From test-taking to test-making: Examining {LLM} authoring of commonsense assessment items.
\newblock In Yaser Al-Onaizan, Mohit Bansal, and Yun-Nung Chen (eds.), \emph{Findings of the Association for Computational Linguistics: EMNLP 2024}, pp.\  5193--5203, Miami, Florida, USA, November 2024. Association for Computational Linguistics.
\newblock \doi{10.18653/v1/2024.findings-emnlp.299}.
\newblock URL \url{https://aclanthology.org/2024.findings-emnlp.299}.

\bibitem[Sharma et~al.(2019)Sharma, Graesser, Nangia, and Evci]{sharma2019naturallanguageunderstandingquora}
Lakshay Sharma, Laura Graesser, Nikita Nangia, and Utku Evci.
\newblock Natural language understanding with the {Q}uora {Q}uestion {P}airs dataset, 2019.
\newblock URL \url{https://arxiv.org/abs/1907.01041}.

\bibitem[Socher et~al.(2013)Socher, Perelygin, Wu, Chuang, Manning, Ng, and Potts]{socher-etal-2013-recursive}
Richard Socher, Alex Perelygin, Jean Wu, Jason Chuang, Christopher~D. Manning, Andrew Ng, and Christopher Potts.
\newblock Recursive deep models for semantic compositionality over a sentiment treebank.
\newblock In David Yarowsky, Timothy Baldwin, Anna Korhonen, Karen Livescu, and Steven Bethard (eds.), \emph{Proceedings of the 2013 Conference on Empirical Methods in Natural Language Processing}, pp.\  1631--1642, Seattle, Washington, USA, October 2013. Association for Computational Linguistics.
\newblock URL \url{https://aclanthology.org/D13-1170/}.

\bibitem[Sun et~al.(2024)Sun, Shaib, and Wallace]{sun2024evaluating}
Jiuding Sun, Chantal Shaib, and Byron~C Wallace.
\newblock Evaluating the zero-shot robustness of instruction-tuned language models.
\newblock In \emph{The Twelfth International Conference on Learning Representations}, 2024.
\newblock URL \url{https://openreview.net/forum?id=g9diuvxN6D}.

\bibitem[Thorne et~al.(2019)Thorne, Vlachos, Christodoulopoulos, and Mittal]{thorne-etal-2019-evaluating}
James Thorne, Andreas Vlachos, Christos Christodoulopoulos, and Arpit Mittal.
\newblock Evaluating adversarial attacks against multiple fact verification systems.
\newblock In Kentaro Inui, Jing Jiang, Vincent Ng, and Xiaojun Wan (eds.), \emph{Proceedings of the 2019 Conference on Empirical Methods in Natural Language Processing and the 9th International Joint Conference on Natural Language Processing (EMNLP-IJCNLP)}, pp.\  2944--2953, Hong Kong, China, November 2019. Association for Computational Linguistics.
\newblock \doi{10.18653/v1/D19-1292}.
\newblock URL \url{https://aclanthology.org/D19-1292/}.

\bibitem[Walkington et~al.(2019)Walkington, Clinton-Lisell, and Sparks]{walkington2019}
Candace Walkington, Virginia Clinton-Lisell, and Anthony Sparks.
\newblock The effect of language modification of mathematics story problems on problem-solving in online homework.
\newblock \emph{Instructional Science}, 47, 10 2019.
\newblock \doi{10.1007/s11251-019-09481-6}.
\newblock URL \url{https://www.jstor.org/stable/48699919}.

\bibitem[Wang et~al.(2018)Wang, Singh, Michael, Hill, Levy, and Bowman]{wang-etal-2018-glue}
Alex Wang, Amanpreet Singh, Julian Michael, Felix Hill, Omer Levy, and Samuel Bowman.
\newblock {GLUE}: A multi-task benchmark and analysis platform for natural language understanding.
\newblock In Tal Linzen, Grzegorz Chrupa{\l}a, and Afra Alishahi (eds.), \emph{Proceedings of the 2018 {EMNLP} Workshop {B}lackbox{NLP}: Analyzing and Interpreting Neural Networks for {NLP}}, pp.\  353--355, Brussels, Belgium, November 2018. Association for Computational Linguistics.
\newblock \doi{10.18653/v1/W18-5446}.
\newblock URL \url{https://aclanthology.org/W18-5446/}.

\bibitem[Wang et~al.(2024{\natexlab{a}})Wang, Wei, Liu, Lin, and Chen]{wang-etal-2024-resilience}
Bin Wang, Chengwei Wei, Zhengyuan Liu, Geyu Lin, and Nancy~F. Chen.
\newblock Resilience of large language models for noisy instructions.
\newblock In Yaser Al-Onaizan, Mohit Bansal, and Yun-Nung Chen (eds.), \emph{Findings of the Association for Computational Linguistics: EMNLP 2024}, pp.\  11939--11950, Miami, Florida, USA, November 2024{\natexlab{a}}. Association for Computational Linguistics.
\newblock \doi{10.18653/v1/2024.findings-emnlp.697}.
\newblock URL \url{https://aclanthology.org/2024.findings-emnlp.697/}.

\bibitem[Wang et~al.(2020)Wang, Pei, Pan, Chen, Wang, and Li]{wang-etal-2020-t3}
Boxin Wang, Hengzhi Pei, Boyuan Pan, Qian Chen, Shuohang Wang, and Bo~Li.
\newblock T3: Tree-autoencoder constrained adversarial text generation for targeted attack.
\newblock In Bonnie Webber, Trevor Cohn, Yulan He, and Yang Liu (eds.), \emph{Proceedings of the 2020 Conference on Empirical Methods in Natural Language Processing (EMNLP)}, pp.\  6134--6150, Online, November 2020. Association for Computational Linguistics.
\newblock \doi{10.18653/v1/2020.emnlp-main.495}.
\newblock URL \url{https://aclanthology.org/2020.emnlp-main.495/}.

\bibitem[Wang et~al.(2022{\natexlab{a}})Wang, Xu, Liu, Cheng, and Li]{wang-etal-2022-semattack}
Boxin Wang, Chejian Xu, Xiangyu Liu, Yu~Cheng, and Bo~Li.
\newblock {S}em{A}ttack: Natural textual attacks via different semantic spaces.
\newblock In Marine Carpuat, Marie-Catherine de~Marneffe, and Ivan~Vladimir Meza~Ruiz (eds.), \emph{Findings of the Association for Computational Linguistics: NAACL 2022}, pp.\  176--205, Seattle, United States, July 2022{\natexlab{a}}. Association for Computational Linguistics.
\newblock \doi{10.18653/v1/2022.findings-naacl.14}.
\newblock URL \url{https://aclanthology.org/2022.findings-naacl.14/}.

\bibitem[Wang et~al.(2022{\natexlab{b}})Wang, Xu, Wang, Gan, Cheng, Gao, Awadallah, and Li]{wang2022adversarialgluemultitaskbenchmark}
Boxin Wang, Chejian Xu, Shuohang Wang, Zhe Gan, Yu~Cheng, Jianfeng Gao, Ahmed~Hassan Awadallah, and Bo~Li.
\newblock Adversarial {GLUE}: A multi-task benchmark for robustness evaluation of language models, 2022{\natexlab{b}}.
\newblock URL \url{https://arxiv.org/abs/2111.02840}.

\bibitem[Wang et~al.(2024{\natexlab{b}})Wang, Hu, Hou, Chen, Zheng, Wang, Yang, Ye, Huang, Geng, Jiao, Zhang, and Xie]{DBLP:journals/debu/0001HH0ZWY0HGJ024}
Jindong Wang, Xixu Hu, Wenxin Hou, Hao Chen, Runkai Zheng, Yidong Wang, Linyi Yang, Wei Ye, Haojun Huang, Xiubo Geng, Binxing Jiao, Yue Zhang, and Xing Xie.
\newblock On the robustness of {C}hat{GPT}: An adversarial and out-of-distribution perspective.
\newblock \emph{IEEE Data Eng. Bull.}, 47\penalty0 (1):\penalty0 48--62, 2024{\natexlab{b}}.
\newblock URL \url{http://sites.computer.org/debull/A24mar/p48.pdf}.

\bibitem[Wang et~al.(2024{\natexlab{c}})Wang, Yang, Huang, Yang, Majumder, and Wei]{wang2024improvingtextembeddingslarge}
Liang Wang, Nan Yang, Xiaolong Huang, Linjun Yang, Rangan Majumder, and Furu Wei.
\newblock Improving text embeddings with large language models, 2024{\natexlab{c}}.
\newblock URL \url{https://arxiv.org/abs/2401.00368}.

\bibitem[Wang et~al.(2022{\natexlab{c}})Wang, Wang, and Yang]{wang-etal-2022-measure}
Xuezhi Wang, Haohan Wang, and Diyi Yang.
\newblock Measure and improve robustness in {NLP} models: A survey.
\newblock In Marine Carpuat, Marie-Catherine de~Marneffe, and Ivan~Vladimir Meza~Ruiz (eds.), \emph{Proceedings of the 2022 Conference of the North American Chapter of the Association for Computational Linguistics: Human Language Technologies}, pp.\  4569--4586, Seattle, United States, July 2022{\natexlab{c}}. Association for Computational Linguistics.
\newblock \doi{10.18653/v1/2022.naacl-main.339}.
\newblock URL \url{https://aclanthology.org/2022.naacl-main.339/}.

\bibitem[Warstadt et~al.(2019)Warstadt, Singh, and Bowman]{warstadt-etal-2019-neural}
Alex Warstadt, Amanpreet Singh, and Samuel~R. Bowman.
\newblock Neural network acceptability judgments.
\newblock \emph{Transactions of the Association for Computational Linguistics}, 7:\penalty0 625--641, 2019.
\newblock \doi{10.1162/tacl_a_00290}.
\newblock URL \url{https://aclanthology.org/Q19-1040/}.

\bibitem[Williams et~al.(2018)Williams, Nangia, and Bowman]{williams-etal-2018-broad}
Adina Williams, Nikita Nangia, and Samuel Bowman.
\newblock A broad-coverage challenge corpus for sentence understanding through inference.
\newblock In Marilyn Walker, Heng Ji, and Amanda Stent (eds.), \emph{Proceedings of the 2018 Conference of the North {A}merican Chapter of the Association for Computational Linguistics: Human Language Technologies, Volume 1 (Long Papers)}, pp.\  1112--1122, New Orleans, Louisiana, June 2018. Association for Computational Linguistics.
\newblock \doi{10.18653/v1/N18-1101}.
\newblock URL \url{https://aclanthology.org/N18-1101/}.

\bibitem[Xu et~al.(2023)Xu, Xie, Li, Wang, Wang, and Li]{10.1145/3593590}
Lingling Xu, Haoran Xie, Zongxi Li, Fu~Lee Wang, Weiming Wang, and Qing Li.
\newblock Contrastive learning models for sentence representations.
\newblock \emph{ACM Trans. Intell. Syst. Technol.}, 14\penalty0 (4), June 2023.
\newblock ISSN 2157-6904.
\newblock \doi{10.1145/3593590}.
\newblock URL \url{https://doi.org/10.1145/3593590}.

\bibitem[Zang et~al.(2020)Zang, Qi, Yang, Liu, Zhang, Liu, and Sun]{zang-etal-2020-word}
Yuan Zang, Fanchao Qi, Chenghao Yang, Zhiyuan Liu, Meng Zhang, Qun Liu, and Maosong Sun.
\newblock Word-level textual adversarial attacking as combinatorial optimization.
\newblock In Dan Jurafsky, Joyce Chai, Natalie Schluter, and Joel Tetreault (eds.), \emph{Proceedings of the 58th Annual Meeting of the Association for Computational Linguistics}, pp.\  6066--6080, Online, July 2020. Association for Computational Linguistics.
\newblock \doi{10.18653/v1/2020.acl-main.540}.
\newblock URL \url{https://aclanthology.org/2020.acl-main.540/}.

\bibitem[Zhang et~al.(2024)Zhang, Hong, Hong, Huang, Wang, Ba, and Ren]{10646716}
Xinyu Zhang, Hanbin Hong, Yuan Hong, Peng Huang, Binghui Wang, Zhongjie Ba, and Kui Ren.
\newblock { Text-{CRS}: A Generalized Certified Robustness Framework against Textual Adversarial Attacks }.
\newblock In \emph{2024 IEEE Symposium on Security and Privacy (SP)}, pp.\  2920--2938, Los Alamitos, CA, USA, May 2024. IEEE Computer Society.
\newblock \doi{10.1109/SP54263.2024.00053}.
\newblock URL \url{https://doi.ieeecomputersociety.org/10.1109/SP54263.2024.00053}.

\bibitem[Zhu et~al.(2024)Zhu, Zhao, Chen, Wang, and Xie]{DBLP:journals/jmlr/ZhuZ00024}
Kaijie Zhu, Qinlin Zhao, Hao Chen, Jindong Wang, and Xing Xie.
\newblock Prompt{B}ench: A unified library for evaluation of large language models.
\newblock \emph{J. Mach. Learn. Res.}, 25:\penalty0 254:1--254:22, 2024.
\newblock URL \url{https://jmlr.org/papers/v25/24-0023.html}.

\end{thebibliography}
\bibliographystyle{iclr2025_conference}

\clearpage

\appendix

\section{Implementation Details}\label{sec:implementation_details}\vspace{5pt}
\subsection{Self-denoising}
For vanilla self-denoising (SD) and iterative self-denoising (SDi), we use greedy decoding with temperature $t=0$.

In Table \ref{tab:hyperparams} we detail the hyperparameters used for training the LoRA modules in the fine-tuned self-denoising pipeline (SFT-SD) on AdvMix. The same hyperparameters are used for both Llama 3 and Flan-T5. Note that we use the hyperparameter combination from \citet{10.5555/3666122.3666563}, since this has been shown to generalise well to a wide range of tasks.\vspace{5pt}

\subsection{Perplexity Smoothing}
All PPL scores in perplexity smoothing are computed using GPT-2 \citep{Radford2019LanguageMA}. 
Candidate substitute words for each \texttt{[MASK]} token are found using DistilRoberta Base\footnote{https://huggingface.co/distilbert/distilroberta-base}. In our experiments, we set $n=10$ (i.e., we mask the top ten most important words). We also set the same beam width $\beta$ as the number of candidates $k$, i.e. $k=\beta=5$, thus performing best-first search.\vspace{5pt}

\subsection{Instruction Ensembling}
For instruction ensembling, we sample $n$ options from the LLM with temperature $t=1$, and set $n=5$. As the classification tasks in our experimental setup are binary, this value of $n$ ensures that it is always possible to take the majority label as the final classification label.\vspace{5pt}

\subsection{Representation Alignment}
We use a siamese model implementation \citep{10.5555/3524938.3525087, 10.1145/3593590, sun2024evaluating} to align the hidden representations of the perturbed instructions to those of the non-perturbed instructions. We align representations at the layer $l$, where $l$ is chosen to be the middle layer of the LLM. For Llama 3 8B Instruct (32 decoder layers), we set $l=16$. For Flan-T5, we take advantage of the encoder-decoder architecture and set $l$ to be last hidden layer of the encoder block. The siamese network is trained on the instruction pairs in AdvMix using LoRA modules (Hu et al., 2022) at the value and query projection layers of the LLM. The LoRA modules are disabled or enabled during each forward pass depending on whether the input consists of unperturbed or perturbed instructions, respectively. Since unperturbed and perturbed instructions may differ in token count, mean pooling is applied to their middle-layer hidden representations before computing the cosine distance loss.

The training hyperparameters---for both the Llama 3 and the Flan-T5 implementation---are shown in Table~\ref{tab:hyperparams} (note that the same hyperparameters are used for training the SFT self-denoising models).

\begin{table}[b!]
    \centering
    \caption{Hyperparameters for training the SFT self-denoising models and the representation alignment network. The hyperparameter combination is the same for both base LLMs (Llama 3 and Flan-T5).}
    \renewcommand\arraystretch{1.2} 
    \setlength{\tabcolsep}{15pt}
    \begin{tabular}{lc}
        \toprule
        \textbf{Hyperparameter} & \textbf{Value}                    \\
        \hline
        LoRA $\alpha$           & $16$                              \\
        LoRA $r$                & $64$                              \\
        LoRA dropout            & $0.1$                             \\
        LoRA modules            & $Q_{\text{proj}},V_{\text{proj}}$ \\
        Learning rate           & $5\mathrm{e}{-5}$                 \\
        Batch size (effective)  & $4$                               \\
        Epochs                  & $10$                              \\
        \bottomrule
    \end{tabular}
    \label{tab:hyperparams}
\end{table}

\clearpage

\section{Self-denoising Meta-prompt and examples}\label{sec:prompt}
\vspace{5pt}
In Prompt \ref{prompt:self-denoise}, we show the meta-prompt and few-shot examples used at both training and inference in the self-denoising pipeline. All examples are extracted from MNLI \citep{williams-etal-2018-broad}, as these achieved the highest validation results across the different datasets, surpassing setups where the exemplar instructions were extracted from multiple diverse tasks. 

\vspace{20pt}
\prompt{prompt:self-denoise}{Meta-prompt and examples for self-denoising}{
\\
Given a sentence which could be perturbed through an adversarial attack, respond with the unperturbed sentence. Do not modify the following words: $\lbrace$excluded\_words$\rbrace$. Do not answer with anything other than the unperturbed sentence.
\\
\\
\\
\\
Uncovering whether the made coupling of condemns revealed entailment, neutral, or contradiction. Cope with ‘entailment’, ‘neutral’, or ‘contradiction’:
\\
\\
Identify whether the given pair of sentences demonstrates entailment, neutral, or contradiction. Answer with ‘entailment’, ‘neutral’, or ‘contradiction’:
\\
\\
\\
Specifies if the made coupling of condemns exposure entailment, neutral, or contradiction. Reacting with ‘entailment’, ‘neutral’, or ‘contradiction’:
\\
\\
Determine if the given pair of sentences displays entailment, neutral, or contradiction. Respond with ‘entailment’, ‘neutral’, or ‘contradiction’:
\\
\\
\\
Can the ratio between the offered penalty be entailment, neutral, or contradiction? Reactions with ‘entailment’, ‘neutral’, or ‘contradiction’:
\\
\\
Does the relationship between the given sentences represent entailment, neutral, or contradiction? Respond with ‘entailment’, ‘neutral’, or ‘contradiction’:
\\
}
\clearpage

\section{Training Data}\label{sec:training_data}

We train the SFT-SD model and the representation alignment pipeline on AdvMix, a custom dataset containing 2,882 pairs of unperturbed and perturbed text sequences.

To create AdvMix, we extract 2,530 pairs of sequences (88\% of the total) from the AdvGLUE benchmark \citep{wang2022adversarialgluemultitaskbenchmark}, which contains the original and perturbed versions of the samples in GLUE \citep{wang-etal-2018-glue}. Our AdvGLUE selection is split among Quora Question Pairs (QQP) \citep{sharma2019naturallanguageunderstandingquora}, Multi-Genre Natural Language Inference (MNLI) \citep{williams-etal-2018-broad}, and Recognizing Textual Entailment (RTE) \citep{10.1007/11736790_9}. Note that none of these datasets are used at test time.

Since AdvGLUE contains general text sequences rather than instructions, we augment AdvMix with 352 samples from PromptBench \citep{DBLP:journals/jmlr/ZhuZ00024}, which contains perturbed and unpreturbed instruction prompts. Again, we are careful to use use datasets that do not appear among our testing data. We extract from Microsoft Research Paraphrase Corpus (MRPC) \citep{dolan-brockett-2005-automatically}.

Note that both AdvGLUE and PromptBench include more than one perturbation for each clean text sequence. To aid the generalization of our train model, we include these different perturbations in AdvMix. We thus source sequences perturbed via four word-level attacks (BERT-Attack \citep{li-etal-2020-bert-attack}, SememePSO \citep{zang-etal-2020-word}, TextBugger \citep{Li_2019}, TextFooler \citep{Jinetal20e}) and seven sentence-level attacks (CheckList \citep{ribeiro-etal-2020-beyond}, AdvFever \citep{thorne-etal-2019-evaluating}, SCPN \citep{iyyer-etal-2018-adversarial}, SemAttack \citep{wang-etal-2022-semattack}, T3 \citep{wang-etal-2020-t3}, StressTest \citep{naik-etal-2018-stress}, and DeepWordBug \citep{8424632}).

\vspace{20pt}

\section{Instructions}\label{sec:instructions}

We use six instruction variants for each dataset, equally split among \textit{task-oriented} and \textit{role-oriented} instructions.

\subsection{CoLA Instructions}
\begin{table}[h!]
    \centering
    \footnotesize
    \begingroup
    \renewcommand{\arraystretch}{1.5}
    \begin{tabular}{|p{\linewidth}|}
        \hline
        \textbf{Task-Oriented}                                                                                                                                                                          \\
        \RaggedRight \texttt{Assess the following sentence and determine if it is grammatically correct. Respond with `acceptable' or `unacceptable':}                                                  \\
        \RaggedRight \texttt{Examine the given sentence and decide if it is grammatically sound. Answer with either `acceptable' or `unacceptable':}                                                    \\
        \RaggedRight \texttt{Analyze the provided sentence and classify its grammatical correctness as `acceptable' or `unacceptable':}                                                                 \\
        \                                                                                                                                                                                               \\
        \textbf{Role-Oriented}                                                                                                                                                                          \\
        \RaggedRight \texttt{In your role as a grammar check tool, assess the following sentence and classify it as `acceptable' if it is grammatically correct or `unacceptable' if it is incorrect:}  \\
        \RaggedRight \texttt{As a grammar identification system, examine the provided sentence and respond with `acceptable' for grammatically correct sentences or `unacceptable' for incorrect ones:} \\
        \RaggedRight \texttt{Functioning as a grammar evaluation tool, analyze the given sentence and decide if it is grammatically correct, responding with `acceptable' or `unacceptable':}           \\
        \hline
    \end{tabular}
    \endgroup
    \label{table:cola_original_prompts}
\end{table}
\clearpage

\subsection{QNLI Instructions}
\begin{table}[!h]
    \centering
    \footnotesize
    \begingroup
    \renewcommand{\arraystretch}{1.5}
    \begin{tabular}{|p{\linewidth}|}
        \hline
        \textbf{Task-Oriented}                                                                                                                                                                   \\
        \RaggedRight \texttt{Given the question and context provided, determine if the answer can be inferred by choosing `entailment' or `not\_entailment'.}                                    \\
        \RaggedRight \texttt{Based on the provided context and question, decide if the information supports the answer by responding with `entailment' or `not\_entailment'.}                    \\
        \RaggedRight \texttt{Please assess if the answer to the question can be derived from the given context by selecting `entailment' or `not\_entailment'.}                                  \\
        \                                                                                                                                                                                        \\
        \textbf{Role-Oriented}                                                                                                                                                                   \\
        \RaggedRight \texttt{As a language expert, assess if the given context entails the answer to the question and respond with `entailment' or `not\_entailment'.}                           \\
        \RaggedRight \texttt{In your role as a semantic evaluator, determine if the provided context justifies the answer to the question and answer with `entailment' or `not\_entailment'.}    \\
        \RaggedRight \texttt{As a textual analyst, examine if the given context logically implies the answer to the question and indicate your decision with `entailment' or `not\_entailment'.} \\
        \hline
    \end{tabular}
    \endgroup
    \label{table:qnli_original_prompts}
\end{table}\vspace{50pt}

\subsection{SST-2 Instructions}
\begin{table}[!ht]
    \centering
    \footnotesize
    \begingroup
    \renewcommand{\arraystretch}{1.5}
    \begin{tabular}{|p{\linewidth}|}
        \hline
        \textbf{Task-Oriented}                                                                                                              \\
        \RaggedRight \texttt{Analyze the tone of this statement and respond with either `positive' or `negative':}                          \\
        \RaggedRight \texttt{Evaluate the sentiment of the given text and classify it as `positive' or `negative':}                         \\
        \RaggedRight \texttt{Please identify the emotional tone of this passage: `positive' or `negative'?}                                 \\
        \                                                                                                                                   \\
        \textbf{Role-Oriented}                                                                                                              \\
        \RaggedRight \texttt{As a sentiment classifier, determine whether the following text is `positive' or `negative'. Please classify:} \\
        \RaggedRight \texttt{In the role of a sentiment analysis tool, respond with `positive' or `negative' to classify this statement:}   \\
        \RaggedRight \texttt{Acting as a sentiment evaluator, identify if the given sentence is `positive' or `negative'. Classify:}        \\
        \hline
    \end{tabular}
    \endgroup
    \label{table:sst2_original_prompts}
\end{table}

\clearpage

\section{Performance with non-perturbed instructions}\label{sec:non-perturbed}
In Table \ref{tab:original_scores} we show the classification accuracy on non-perturbed instructions for each method. Scores are averaged across datasets (CoLA, QNLI, SST-2), underlying LLMs (Llama 3, Flan-T5) and instruction variants (six variants for each dataset). Note that the accuracy scores obtained by the augmented pipelines are within only 1\% of that achieved using the base model implementation.
\vspace{10pt}

\begin{table}[h!]
    \caption{Accuracy scores for each method, averaged across datasets and LLMs.}
    \centering
    \renewcommand\arraystretch{1.2} 
    \setlength{\tabcolsep}{15pt}
    \begin{tabular}{lcc}
        \toprule
        Method  & Avg. performance                  \\
        \midrule

        Base LLM & 80.1 \\
        \cmidrule(lr){1-3}        
        PPL smoothing & 79.3 \\
        Instruction ensembling & 80.0 \\  
        Representation alignment & 79.8 \\
        SD & 80.0 \\
        SDi & 80.0 \\
        SFT-SDi  & 79.1 \\
        \bottomrule
    \end{tabular}

    \label{tab:original_scores}
\end{table}

\end{document}